\documentclass[10pt,twocolumn,letterpaper]{article}

\usepackage{cvpr}
\usepackage{times}
\usepackage{epsfig}
\usepackage{graphicx}
\usepackage{amsmath}
\usepackage{amssymb}

\usepackage{multirow, makecell}
\usepackage{lineno}
\usepackage{enumerate}
\usepackage{amsfonts}
\usepackage{gensymb}
\usepackage{bm}
\usepackage{diagbox}
\usepackage{tabularx}
\usepackage{rotating}
\newcolumntype{L}[1]{>{\raggedright\let\newline\\\arraybackslash\hspace{0pt}}m{#1}}
\newcolumntype{C}[1]{>{\centering\arraybackslash}m{#1}}
\newcolumntype{R}[1]{>{\raggedleft\let\newline\\\arraybackslash\hspace{0pt}}m{#1}}
\usepackage{xmpmulti}
\usepackage{arydshln}

\usepackage[breaklinks=true,bookmarks=false]{hyperref}

\cvprfinalcopy % *** Uncomment this line for the final submission

\def\cvprPaperID{5950} % *** Enter the CVPR Paper ID here

\ifcvprfinal\fi
\begin{document}

%%%%%%%%% TITLE
\title{Learning to Discriminate Information for Online Action Detection}

\author{Hyunjun~Eun\textsuperscript{1,4}\thanks{Work done during a Ph.D. student at KAIST.}
\and Jinyoung~Moon\textsuperscript{2} 
\and Jongyoul~Park\textsuperscript{2} 
\and Chanho~Jung\textsuperscript{3} 
\and Changick~Kim\textsuperscript{1} 
\and\textsuperscript{1}Korea Advanced Institute of Science and Technology (KAIST)\\
\textsuperscript{2}Electronics and Telecommunications Research Institute (ETRI)\\
\textsuperscript{3}Hanbat National University
\hspace{1cm} \textsuperscript{4}SK Telecom\\
}

\maketitle
\thispagestyle{empty}

%%%%%%%%% ABSTRACT
\begin{abstract}
From a streaming video, online action detection aims to identify actions in the present.
For this task, previous methods use recurrent networks to model the temporal sequence of current action frames.
However, these methods overlook the fact that an input image sequence includes background and irrelevant actions as well as the action of interest.
For online action detection, in this paper, we propose a novel recurrent unit to explicitly discriminate the information relevant to an ongoing action from others.
Our unit, named Information Discrimination Unit (IDU), decides whether to accumulate input information based on its relevance to the current action.
This enables our recurrent network with IDU to learn a more discriminative representation for identifying ongoing actions.
In experiments on two benchmark datasets, TVSeries and THUMOS-14, the proposed method outperforms state-of-the-art methods by a significant margin.
Moreover, we demonstrate the effectiveness of our recurrent unit by conducting comprehensive ablation studies.
\end{abstract}

%%%%%%%%% BODY TEXT
\section{Introduction}
Temporal action detection \cite{chao2018cvpr,liu2019cvpr,xu2017iccv,zhang2019aaai,zhao2017iccv} has been widely studied in an offline setting, which allows making a decision for the detection after fully observing a long untrimmed video.
This is called offline action detection.
In contrast, online action detection aims to identify ongoing actions from streaming videos, at every moment in time.
This task is useful for many real-world applications (e.g., autonomous driving \cite{kim2019cvpr}, robot assistants \cite{koppula2013iros}, and surveillance systems \cite{iwashita2013bmvc, shu2015cvpr}).

\begin{figure}[t!]
\centering{\includegraphics[width=.99\linewidth]{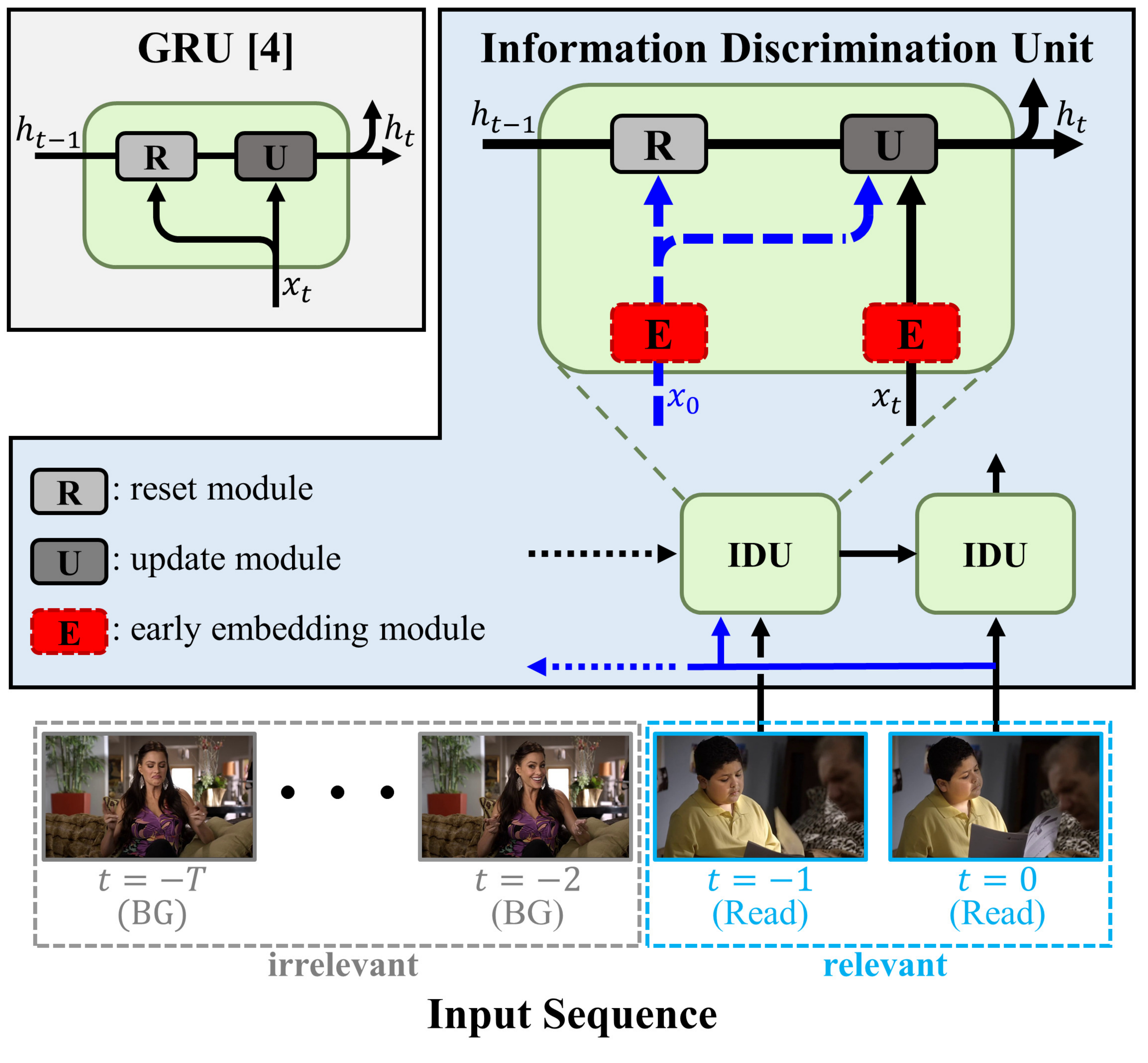}}
\caption{\label{fig1}Comparison between GRU \cite{cho2014emnlp} and Information Discrimination Unit (IDU) which is proposed for our online action detection system.
Our IDU extends GRU with two novel components, a mechanism utilizing current information (blue lines) and an early embedding module (red dash boxes).
First, reset and update modules in our IDU additionally takes the current information (i.e., $x_0$), which enables to consider whether the past information (i.e, $h_{t-1}$ and $x_t$) are relevant to an ongoing action such as $x_0$.
Second, the early embedding module is introduced to consider the relation between high-level features for both information.
}
\end{figure}

Recent methods \cite{gao22017bmvc,xu2019iccv} for online action detection mostly employ recurrent neural networks (RNNs) with recurrent units (e.g., long short-term memory (LSTM) \cite{hochreiter1997nc} and gated recurrent unit (GRU) \cite{cho2014emnlp}) for modeling the temporal sequence of an ongoing action.
They introduce additional modules to learn a discriminative representation.
However, these methods overlook the fact that the given input video contains not only the ongoing action but irrelevant actions and background.
Specifically, the recurrent unit accumulates the input information without explicitly considering its relevance to the current action, and thus the learned representation would be less discriminative.
Note that, in the task of detecting actions online, ignoring such a characteristic of streaming videos makes the problem more challenging \cite{geest2016eccv}.

In this paper, we investigate on the question of \textit{how RNNs can learn to explicitly discriminate relevant information from irrelevant information for detecting actions in the present}.
To this end, we propose a novel recurrent unit that extends GRU \cite{cho2014emnlp} with a mechanism utilizing current information and an early embedding module (see Fig. \ref{fig1}).
We name our recurrent unit Information Discrimination Unit (IDU). 
Specifically, our IDU models the relation between an ongoing action and past information (i.e., $x_t$ and $h_{t-1}$) by additionally taking current information (i.e., $x_0$) at every time step.
We further introduce the early embedding module to more effectively model the relation.
By adopting action classes and feature distances as supervisions, our embedding module learns the features for the current and past information describing actions in a high level.
Based on IDU, our Information Discrimination Network (IDN) effectively determines whether to use input information in terms of its relevance to the current action.
This enables the network to learn a more discriminative representation for detecting ongoing actions.
We perform extensive experiments on two benchmark datasets, where our IDN achieves state-of-the-art performances of 86.1\% mcAP and 60.3\% mAP on TVSeries \cite{geest2016eccv} and THUMOS-14 \cite{jiang2015url}, respectively.
These performances significantly outperform TRN \cite{xu2019iccv}, the previous best performer, by 2.4\% mcAP and 13.1\% mAP on TVSeries and THUMOS-14, respectively.

Our contributions are summarized as follows:
\renewcommand\labelitemi{\tiny$\bullet$}
\begin{itemize}
\item Different from previous methods, we investigate on how recurrent units can explicitly discriminate relevant information from irrelevant information for online action detection.
\item We introduce a novel recurrent unit, IDU, with a mechanism using current information at every time step and an early embedding module to effectively model the relevance of input information to an ongoing action.
\item We demonstrate that our IDN significantly outperforms state-of-the-arts in extensive experiments on two benchmark datasets.
\end{itemize}

\section{Related Work}
\textbf{Offline Action Detection.}
The goal of offline action detection is to detect the start and end times of action instances from fully observed long untrimmed videos.
Most methods \cite{chao2018cvpr,shou2016cvpr,zhao2017iccv} consist of two steps including action proposal generation and action classification.
SSN \cite{zhao2017iccv} first evaluates actionness scores for temporal locations to generate temporal intervals.
Then, these intervals are classified by modeling the temporal structures and completeness of action instances.
TAL-Net \cite{chao2018cvpr} including the proposal generation and classification networks is the extended version of Faster R-CNN \cite{ren2015nips} for offline action detection.
This method changes receptive field alignment, the range of receptive fields, and feature fusion to fit the action detection.
%Several methods \cite{gao2018eccv,lin2019iccv,lin2018eccv,liu2019cvpr} have been also studied for temporal action proposal generation.
%They show large performance gains by combining promising action proposals with existing action classifiers \cite{simonyan2014nips,simonyan2015iclr}.
Other methods \cite{donahue2015cvpr,yeung2018ijcv} with LSTM have been also studied for per-frame prediction.

\textbf{Early Action Prediction.}
This task is similar to online action detection but focuses on recognizing actions from the partially observed videos.
Hoai and la Torre \cite{hoai2014ijcv} introduced a maximum margin framework with the extended structured SVM \cite{tsochantaridis2005jmlr} to accommodate sequential data.
Cai \emph{et al.} \cite{cai2019aaai} proposed to transfer the action knowledge learned from full actions for modeling partial actions.

\textbf{Online Action Detection.}
Given a streaming video, online action detection aims to identify actions as soon as each video frame arrives, without observing future video frames.
Geest \emph{et al.} \cite{geest2016eccv} introduced a new large dataset, TVSeries, for online action detection.
They also analyzed and compared several baseline methods on the TVseries dataset.
In \cite{geest2018wacv}, a two-stream feedback network with LSTM is proposed to individually perform the interpretation of the features and the
modeling of the temporal dependencies.
Gao, Yang, and Nevatia \cite{gao22017bmvc} proposed an encoder-decoder network with a reinforcement module, of which the reward function encourages the network to make correct decisions as early as possible.
TRN \cite{xu2019iccv} predicts future information and utilizes the predicted future as well as the past and current information together for detecting a current action.

Aforementioned methods \cite{geest2016eccv,gao22017bmvc,xu2019iccv} for online action detection adopt RNNs to model a current action sequence.
However, the RNN units such as LSTM \cite{hochreiter1997nc} and GRU \cite{cho2014emnlp} operate without explicitly considering whether input information is relevant to the ongoing action.
Therefore, the current action sequence is modeled based on both relevant and irrelevant information, which results in a less discriminative representation.

\begin{figure*}[t!]
\centering
\begin{minipage}[t]{0.499\linewidth}
{\centering{\includegraphics[width=.99\linewidth]{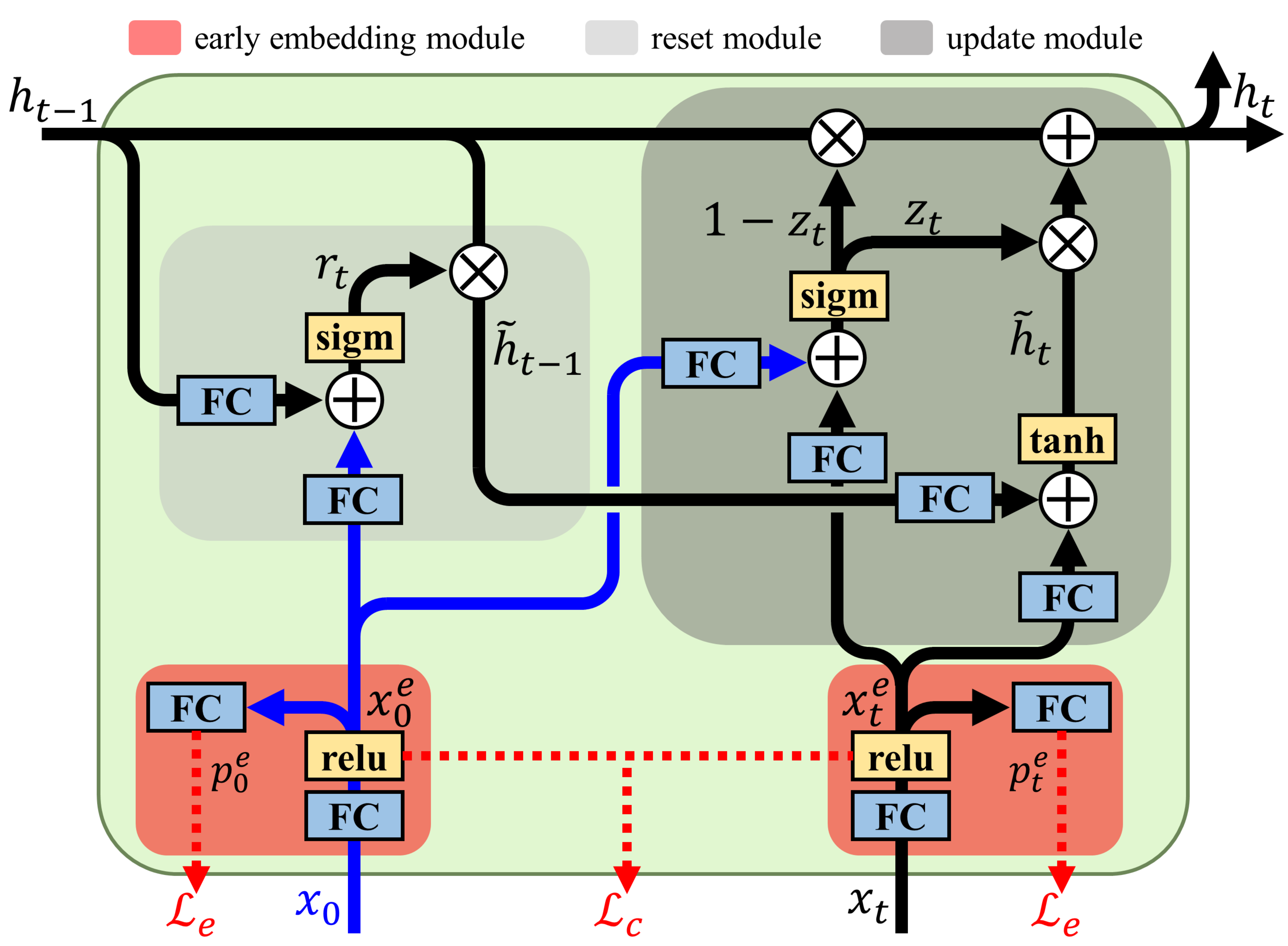}}}
\centering{\small{(a) Information Discrimination Unit (IDU)}}
\end{minipage}
\begin{minipage}[t]{0.495\linewidth}
\centering{\includegraphics[width=.95\linewidth]{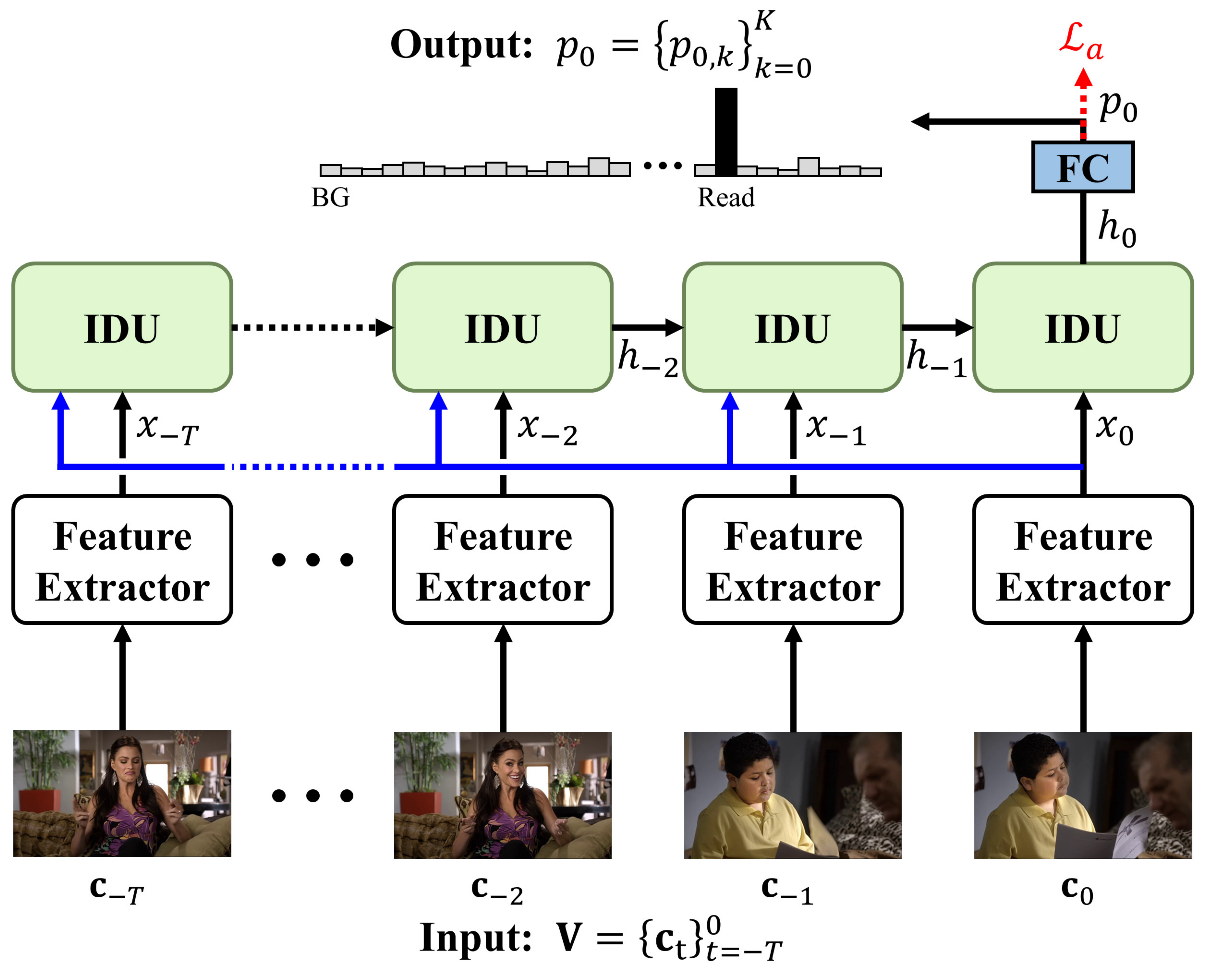}} 
\centering{\small{(b) Information Discrimination Network (IDN)}}
\end{minipage}
\vspace{0.05cm}
\caption{\label{fig2}Illustration of our Information Discrimination Unit (IDU) and Information Discrimination Network (IDN).
(a) Our IDU extends GRU with two new components, a mechanism using current information (i.e., $x_0$) (blue lines) and an early embedding module (red boxes). The first encourages reset and update modules to model the relation between past information (i.e., $h_{t-1}$ and $x_t$) and an ongoing action. The second enables to effectively model the relation between high-level features for the input information.
(b) Given an input streaming video $\mathbf{V}=\{\mathbf{c}_t\}_{t=-T}^0$ consisting of sequential chunks, IDN models a current action sequence and outputs the probability distribution $p_0$ of the current action over $K$ action classes and background.
}
\end{figure*}

\section{Preliminary: Gated Recurrent Units}
We first analyze GRU \cite{cho2014emnlp} to compare differences between the proposed IDU and GRU.
GRU is one of the recurrent units, which is much simpler than LSTM.
Two main components of GRU are reset and update gates.

The reset gate $r_t$ is computed based on a previous hidden state $h_{t-1}$ and an input $x_t$ as follows:
\begin{eqnarray}
r_t = \sigma (\textbf{W}_{hr}h_{t-1} + \textbf{W}_{xr}x_t),
\end{eqnarray}
where $\textbf{W}_{hr}$ and $\textbf{W}_{xr}$ are parameters to be trained and $\sigma$ is the logistic sigmoid function.
Then, the reset gate determines whether a previous hidden state $h_{t-1}$ is ignored as
\begin{eqnarray}
\tilde{h}_{t-1} = r_t \odot h_{t-1},
\end{eqnarray}
where $\tilde{h}_{t-1}$ is a new previous hidden state. 

Similar to $r_t$, the update gate $z_t$ is also computed based on $h_{t-1}$ and $x_t$ as
\begin{eqnarray}
z_t = \sigma (\textbf{W}_{xz}x_t + \textbf{W}_{hz}h_{t-1}),
\end{eqnarray}
where $\textbf{W}_{xz}$ and $\textbf{W}_{hz}$ are learnable parameters.
The update gate decides whether a hidden state $h_t$ is updated with a new hidden state $\tilde{h}_t$ as follows:
\begin{eqnarray}
h_t = (1-z_t) \odot h_{t-1} + z_t \odot \tilde{h}_t,
\end{eqnarray}
where
\begin{eqnarray}
\tilde{h}_t = \eta (\textbf{W}_{x\tilde{h}}x_t + \textbf{W}_{\tilde{h}\tilde{h}}\tilde{h}_{t-1}).
\end{eqnarray}
Here $\textbf{W}_{x\tilde{h}}$ and $\textbf{W}_{\tilde{h}\tilde{h}}$ are trainable parameters and $\eta$ is the tangent hyperbolic function.

Based on reset and update gates, GRU effectively drops and accumulates information to learn a compact representation.
However, there are limitations when we applied GRU to online action detection as below:

First, the past information including $x_t$ and $h_{t-1}$ directly effects the decision of the reset and update gates.
For online action detection, the relevant information to be accumulated is the information related to a current action.
Thus, it is advantageous to make a decision based on the relation between the past information and the current action instead.
To this end, we reformulate the computations of the reset and update gates by additionally taking the current information (i.e., $x_0$) as input. 

This enables the reset and update gates to drop the irrelevant information and accumulate the relevant information regarding the ongoing action.
Second, it is implicitly considered that the input features that the reset and update gates use represent valuable information.
We augment GRU with an early embedding module with supervisions, action classes and feature distances, so that the input features explicitly describe actions.
By optimizing features for the target task and dataset, our early embedding module also lets the reset and update gates focus on accumulating the relevant information along with the recurrent steps.

\section{Approach}
We present the schematic view of our IDU and the framework of IDN in Fig. \ref{fig2}.
We first describe our IDU in details and then explain on IDN for online action detection.

\subsection{Information Discrimination Units}
Our IDU extends GRU with two new components, a mechanism utilizing current information (i.e., $x_0$) and an early embedding module.
We explain IDU with early embedding, reset, and update modules, which takes a previous hidden state $h_{t-1}$, the features at each time $x_t$, and the features at current time $x_0$ as input and outputs a hidden state $h_t$ (see Fig. \ref{fig2}.(a)).

\textbf{Early Embedding Module.}
Our early embedding module individually processes the features at each time $x_t$ and the features at current time $x_0$ and outputs embedded features $x_t^e$ and $x_0^e$ as follows:
\begin{eqnarray}
x_t^e = \zeta(\textbf{W}_{xe}x_t), \\
x_0^e = \zeta(\textbf{W}_{xe}x_0),
\end{eqnarray}
where $\bm{W}_{xe}$ is a weight matrix and $\zeta$ is the ReLU \cite{nari2010icml} activation function.
Note that we share $\bm{W}_{xe}$ for $x_t$ and $x_0$.
We omit a bias term for simplicity.

To encourage $x_t^e$ and $x_0^e$ to represent specific actions, we introduce two supervisions, action classes and feature distances.
First, we process $x_t^e$ and $x_0^e$ to obtain probability distributions $p_t^e$ and $p_0^e$ over $K$ action classes and background:
\begin{eqnarray}
p_t^e = \xi(\textbf{W}_{ep}x_t^e), \\
p_0^e = \xi(\textbf{W}_{ep}x_0^e),
\end{eqnarray}
where $\bm{W}_{ep}$ is a shared weight matrix to be learned and $\xi$ is the softmax function.
We design a classification loss ${\cal{L}}_e$ by adopting the multi-class cross-entropy loss as
\begin{eqnarray}
{\cal{L}}_e = -\sum_{k=0}^K \left( \ y_{t,k} \text{log}(p_{t,k}^e) + y_{0,k} \text{log}(p_{0,k}^e) \right),
\end{eqnarray}
where $y_{t,k}$ and $y_{0,k}$ are ground truth labels.
Second, we use the contrastive loss \cite{chopra2005cvpr,hadsell2006cvpr} proposed to learn an embedding representation by preserving the distance between similar data points close and dissimilar data points far on the embedding space in metric learning \cite{sohn2016nips}.
By using $x_t^e$ and $x_0^e$ as a pair, we design our contrastive loss ${\cal{L}}_c$ as
\begin{eqnarray}
\label{eq11}
\begin{aligned}
{\cal{L}}_c = & \textbf{1} \{ y_t=y_0 \} D^2(x_t^e, x_0^e) \\
& + \textbf{1} \{ y_t \neq y_0 \} \text{max}(0, m-D^2(x_t^e, x_0^e)),
\end{aligned}
\end{eqnarray}
where $D^2(a,b)$ is the squared Euclidean distance and $m$ is a margin parameter.

We train our embedding module with ${\cal{L}}_e$ and ${\cal{L}}_c$, which provides more representative features for actions.
More details on training will be provided in Section 4.2.

\textbf{Reset Module.}
Our reset module takes the previous hidden state $h_{t-1}$ and the embedded features $x_0^e$ to compute a reset gate $r_t$ as 
\begin{eqnarray}
\label{eq12}
r_t = \sigma(\textbf{W}_{hr}h_{t-1} + \textbf{W}_{x_0r}x_0^e),
\end{eqnarray}
where $\textbf{W}_{hr}$ and $\textbf{W}_{x_0r}$ are weight matrices which are learned.
We define $\sigma$ as the logistic sigmoid function same as GRU.
We then obtain a new previous hidden state $\tilde{h}_{t-1}$ as follows:
\begin{eqnarray}
\tilde{h}_{t-1} = r_t \odot h_{t-1}.
\end{eqnarray}

Different from GRU, we compute the reset gate $r_t$ based on $h_{t-1}$ and $x_0^e$.
This enables our reset gate to effectively drop or take the past information according to its relevance to an ongoing action.

\textbf{Update Module.}
Our update module adopts the embedded features $x_t^e$ and $x_0^e$ to compute an update gate $z_t$ as follows:
\begin{eqnarray}
\label{eq14}
z_t = \sigma(\textbf{W}_{x_tz}x_t^e + \textbf{W}_{x_0z}x_0^e),
\end{eqnarray}
where $\textbf{W}_{x_tz}$ and $\textbf{W}_{x_0z}$ are trainable parameters.
Then, a hidden state $h_t$ is computed as follows:
\begin{eqnarray}
h_t = (1-z_t) \odot h_{t-1} + z_t \odot \tilde{h}_t,
\end{eqnarray}
where 
\begin{eqnarray}
\label{eq16}
\tilde{h}_t = \eta (\textbf{W}_{x_t\tilde{h}}x_t^e + \textbf{W}_{\tilde{h}\tilde{h}}\tilde{h}_{t-1}).
\end{eqnarray}
Here $\tilde{h}_t$ is a new hidden state and $\eta$ is the tangent hyperbolic function.
$\textbf{W}_{x_t\tilde{h}}$ and $\textbf{W}_{\tilde{h}\tilde{h}}$ are trainable parameters.

There are two differences between the update modules of our IDU and GRU. 
The first difference is that our update gate is computed based on $x_t^e$ and $x_0^e$.
This allows the update gate to consider whether $x_t^e$ is relevant to an ongoing action.
Second, our update gate uses the embedded features which are more representative in terms of specific actions.

\subsection{Information Discrimination Network}
In this section, we explain our recurrent network, called IDN, for online action detection (see Fig. \ref{fig2}.(b)).

\textbf{Problem Setting.}
To formulate the online action detection problem, we follow the same setting as in previous methods \cite{gao22017bmvc,xu2019iccv}.
Given a streaming video $\textbf{V}=\{ \textbf{c}_t \}_{t=-T}^0$ including current and $T$ past chunks as input, our IDN outputs a probability distribution $p_0= \{ p_{0,k}\}_{k=0}^K$ of a current action over $K$ action classes and background.
Here we define a chunk $c= \{ I_n \}_{n=1}^N$ as the set of $N$ consecutive frames.
$I_n$ indicates the $n$th frame.

\textbf{Feature Extractor.}
We use TSN \cite{wang2016eccv} as a feature extractor.
TSN takes an individual chunk $\textbf{c}_t$ as input and outputs an appearance feature vector $x_t^a$ and a motion feature vector $x_t^m$.
We concatenate $x_t^a \in \mathbb{R}^{d_{a}}$ and $x_t^m \in \mathbb{R}^{d_{m}}$ into a two-stream feature vector $x_t=[x_t^a, x_t^m]\in \mathbb{R}^{d_{x}}$.
Here $d_x$ equals to $d_a + d_m$.
After that, we sequentially feed $x_t$ and $x_0$ into our IDU.

\textbf{Training.}
We feed the hidden state $h_0$ at current time into a fully connected layer to obtain the final probability distribution $p_0$ of an ongoing action as follows:
\begin{eqnarray}
p_0^e = \xi(\textbf{W}_{hp}h_0),
\end{eqnarray}
where $\textbf{W}_{hp}$ is a trainable matrix and $\xi$ is the softmax function.

We define a classification loss ${\cal{L}}_a$ for a current action by employing the standard cross-entropy loss as
\begin{eqnarray}
{\cal{L}}_a = -\sum_{t=-T}^0 \sum_{k=0}^K \ y_{t,k} \text{log}(p_{t,k}),
\end{eqnarray}
where $y_{t,k}$ are the ground truth labels for the $t$th time step.
We train our IDN by jointly optimizing ${\cal{L}}_a$, ${\cal{L}}_e$, and ${\cal{L}}_c$ by designing a multi-task loss ${\cal{L}}$ as follows:
\begin{eqnarray}
\label{eq18}
{\cal{L}} = {\cal{L}}_a + \alpha ( {\cal{L}}_e + {\cal{L}}_c ),
\end{eqnarray}
where $\alpha$ is a balance parameter.

\section{Experiments}
In this section, we evaluate the proposed method on two benchmark datasets, TVSeries \cite{geest2016eccv} and THUMOS-14 \cite{jiang2015url}.
We first demonstrate the effectiveness of our IDU by conducting comprehensive ablation studies.
We then report comparison results among our IDN and the state-of-the-art methods for online action detection.

\subsection{Datasets}
\textbf{TVSeries \cite{geest2016eccv}.} This dataset includes 27 untrimmed videos on six popular TV series, divided into 13, 7, and 7 videos for training, validation, and test, respectively.
Each video contains a single episode, approximately 20 minutes or 40 minutes long.
The dataset is temporally annotated with 30 realistic actions (e.g., open door, read, eat, \textit{etc}).
The TVSeries dataset is challenging due to diverse undefined actions, multiple actors, heavy occlusions, and a large proportion of non-action frames.

\textbf{THUMOS-14 \cite{jiang2015url}.} The THUMOS-14 dataset consists of 200 and 213 untrimmed videos for validation and test sets, respectively.
This dataset has temporal annotations with 20 sports actions (e.g., diving, shot put, billiards, \textit{etc}).
Each video includes 15.8 action instances and 71$\%$ background on average.
As done in \cite{gao22017bmvc,xu2019iccv}, we used the validation set for training and the test set for evaluation.

\subsection{Evaluation Metrics}
For evaluating performance in online action detection, existing methods \cite{geest2016eccv,gao22017bmvc,xu2019iccv} measure mean average precision (mAP) and mean calibrated average precision (mcAP) \cite{geest2016eccv} in a frame-level.
Both metrics are computed in two steps: 1) calculating the average precision over all frames for each action class and 2) averaging the average precision values over all action classes.

\textbf{mean Average Precision (mAP).}
On each action class, all frames are first sorted in descending order of their probabilities.
The average precision of the $k$th class over all frames is then calculated based on the precision at cut-off $i$ (i.e., on the $i$ sorted frames).
The final mAP is defined as the mean of the AP values over all action classes.
%On each action class, all frames are first sorted in descending order of their probabilities.
%The precision at cut-off $i$ (i.e., on the $i$ sorted frames) is then computed as
%\begin{eqnarray}
%\text{Prec}(i) = \frac{\text{TP}(i)}{\text{TP}(i)+\text{FP}(i)},
%\end{eqnarray}
%where TP$( \cdot )$ and FP$( \cdot )$ are the numbers of true positive and false positive frames, respectively.
%Based on Eq. (20), the average precision of the $k$th class over all frames is computed as follows:
%\begin{eqnarray}
%\text{AP}_k=\frac{\sum_i \text{Prec}(i)\textbf{1}(i)}{N_P},
%\end{eqnarray}
%where $N_P$ is the total number of positive frames.
%$\textbf{1}(i)$ equals to 1 if the frame $i$ is a true positive, otherwise 0.
%The final mAP is then defined as the mean of the AP values over all action classes.

\textbf{mean calibrated Average Precision (mcAP).}
It is difficult to compare two different classes in terms of the AP values when the ratios of positive frames versus negative frames for these classes are different.
To address this problem, Geest \emph{et al.} \cite{geest2016eccv} propose the calibrated precision as
\begin{eqnarray}
\text{cPrec}(i) = \frac{w\text{TP}(i)}{w\text{TP}(i)+\text{FP}(i)},
\end{eqnarray}
where $w$ is a ratio between negative frames and positive frames.
Similar to the AP, the calibrated average precision of the $k$th class over all frames is computed as
\begin{eqnarray}
\text{cAP}_k=\frac{\sum_i \text{cPrec}(i)\textbf{1}(i)}{N_P}.
\end{eqnarray}
Then, the mcAP is obtained by averaging the cAP values over all action classes.

\subsection{Implementation Details}

\textbf{Problem Setting.}
We use the same setting as used in state-of-the-art methods \cite{gao22017bmvc,xu2019iccv}.
On both TVSeries \cite{geest2016eccv} and THUMOS-14 \cite{jiang2015url} datasets, we extract video frames at $24$ fps and set the number of frames in each chunk $N$ to $6$.
We use $16$ chunks (i.e., $T=15$), which are $4$ seconds long, for the input of IDN.

\textbf{Feature Extractor.}
We use a two-stream network as a features extractor.
In the two-stream network, one stream encodes appearance information by taking the center frame of a chunk as input, while another stream encodes motion information by processing an optical flow stack computed from an input chunk.
Among several two-stream networks, we employ the TSN model \cite{wang2016eccv} pretrained on the ActivityNet-v1.3 dataset \cite{heilbron2015cvpr}.
Note that this TSN is the same feature extractor as used in state-of-the-art methods \cite{gao22017bmvc, xu2019iccv}.
The TSN model consists of ResNet-200 \cite{he2016cvpr} for an appearance network and BN-Inception \cite{ioffe2015arxiv} for a motion network.
We use the outputs of the {\fontfamily{qcr}\selectfont Flatten\_673} layer in ResNet-200 and the {\fontfamily{qcr}\selectfont global\_pool} layer in BN-Inception as the appearance features $x_t^a$ and motion features $x_t^m$, respectively.
The dimensions of $x_t^a$ and $x_t^m$ are $d_a=2048$ and $d_m=1024$, respectively, and $d_x$ equals to $3072$.

\textbf{IDN Architecture.}
Table 1 provides the specifications of IDN considered in our experiments.
In the early embedding module, we set the number of the hidden units for $\textbf{W}_{xe}$ to $512$.
In the reset module, both weights $\textbf{W}_{hr}$ and $\textbf{W}_{x_ar}$ have 512 hidden units.
In the update module, we use 512 hidden units for $\textbf{W}_{x_tz}$, $\textbf{W}_{x_0z}$, $\textbf{W}_{x_t\tilde{h}}$, and $\textbf{W}_{\tilde{h}\tilde{h}}$.
According to the number of action classes, we set $K+1$ to 31 for TVSeries and 21 for THUMOS-14.

\begin{table}[t]
\centering
\begin{tabular}{C{2.6cm}|C{.8cm}|C{1.2cm}|C{1.9cm}}\hline 
Module & Type & Weight & Size  \\ \hline\hline
\multirow{2}{*}{\shortstack{Early Embedding\\Module}} & FC & $\textbf{W}_{xe}$ & $d_x\times512$  \\
 & FC & $\textbf{W}_{ep}$ & $512\times (K+1)$ \\ \hline
\multirow{2}{*}{\shortstack{Reset\\Module}} & FC & $\textbf{W}_{hr}$ & $512\times512$ \\
 & FC & $\textbf{W}_{x_ar}$ & $512\times512$ \\ \hline
\multirow{4}{*}{\shortstack{Update\\Module}} & FC & $\textbf{W}_{x_tz}$ & $512\times512$ \\
 & FC & $\textbf{W}_{x_0z}$ & $512\times512$ \\
 & FC & $\textbf{W}_{x_t\tilde{h}}$ & $512\times 512$ \\
 & FC & $\textbf{W}_{\tilde{h}\tilde{h}}$ & $512\times 512$ \\ \hline
Classification & FC & $\textbf{W}_{hp}$ & $512\times (K+1)$  \\ \hline
\end{tabular}
\centering
\caption{\label{tab1}Specifications of our IDN. $d_x$ is the dimension of the two-stream feature vector $x_t$ and $K+1$ is the number of action and background classes.}
\end{table}

\textbf{IDN Training.}
To train our IDN, we use a stochastic gradient descent optimizer with the learning rate of 0.01 for both THUMOS-14 and TVSeries datasets.
We set batch size to 128 and balance the numbers of action and background samples in terms of the class of $\textbf{c}_0$.
We empirically set the margin parameter $m$ in Eq. (\ref{eq11}) to $1.0$ and the balance parameter $\alpha$ in Eq. (\ref{eq18}) to $0.3$.

\subsection{Ablation Study}
We evaluate RNNs with the simple unit, LSTM \cite{hochreiter1997nc}, and GRU \cite{cho2014emnlp}.
We name these networks RNN-Simple, RNN-LSTM, and RNN-GRU, respectively.
Although many methods \cite{geest2016eccv,gao22017bmvc,xu2019iccv} report the performances of these networks as baselines, we evaluate them in our setting to clearly confirm the effectiveness of our IDU.

In addition, we individually add IDU components to GRU as a baseline for analyzing their effectiveness:\\
\textbf{Baseline+CI:} We add a mechanism using current information to GRU in computing reset and update gates. 
Specifically, we replace Eq. (1) for $r_t$ with
\begin{eqnarray}
r_t = \sigma (\textbf{W}_{hr}h_{t-1} + \textbf{W}_{x_0r}x_0)
\end{eqnarray}
and Eq. (3) for $z_t$ with
\begin{eqnarray}
z_t = \sigma (\textbf{W}_{x_tz}x_t + \textbf{W}_{x_0z}x_0),
\end{eqnarray}
where $\textbf{W}_{hr}$, $\textbf{W}_{x_0r}$, $\textbf{W}_{x_tz}$, and $\textbf{W}_{x_0z}$ are trainable parameters.
We construct a recurrent network with this modified unit.
\\
\textbf{Baseline+CI+EE (IDN):} We incorporate our main components, a mechanism utilizing current information and an early embedding module, into GRU, which is our IDU.
These components enable reset and update gates to effectively model the relation between an ongoing action and input information at every time step.
Specifically, Eq. (12) and Eq. (14) are substituted for Eq. (1) and Eq. (3), respectively.
We design a recurrent network with our IDU, which is the proposed IDN. \\

In Table \ref{tab2}, we report the performances of five networks on the TVSeries dataset \cite{geest2016eccv}.
Among RNN-Simple, RNN-LSTM, and RNN-GRU, RNN-GRU results in the highest mcAP of 81.3\%.
By comparing RNN-GRU (Baseline) with Baseline+CI, we first analyze the effect of using $x_0$ in calculating reset and update gates.
This component enables the gates to decide whether input information at each time is relevant to a current action.
As a result, Baseline-CI achieves the performance gain of 2.1\% mcAP, which demonstrates the effectiveness of using $x_0$.
Next, we observe that adding the early embedding module improves the performance by 1.3\% mcAP from the comparison between Baseline+CI and Baseline+CI+EE (IDN).
Note that our IDN achieves mcAP of 84.7\% with a performance gain of 3.4\% mcAP compared with Baseline.

\begin{table}[t]
\centering
\begin{tabular}{C{5.5cm}|C{2.cm}}\hline
Method & mcAP (\%) \\ \hline\hline
RNN-Simple & 79.9\\
RNN-LSTM & 80.9 \\
RNN-GRU (Baseline) & 81.3 \\ \cdashline{1-2}
Baseline+CI & 83.4 \\
Baseline+CI+EE (IDN) & \bf{84.7} \\ \hline
\end{tabular}
\centering
\caption{\label{tab2}Ablation study of the effectiveness of our proposed components on TVSeries \cite{geest2016eccv}. CI and EE indicate additionally using the current information and early embedding input information, respectively.}
\end{table}

\begin{table}[t]
\centering
\begin{tabular}{C{5.5cm}|C{2.cm}}\hline 
Method& mAP (\%) \\ \hline\hline
RNN-Simple & 45.5 \\
RNN-LSTM & 46.3 \\ 
RNN-GRU (Baseline) & 46.7 \\ \cdashline{1-2}
Baseline+CI & 48.6 \\
Baseline+CI+EE (IDN) & \bf{50.0} \\ \hline
\end{tabular}
\centering
\caption{\label{tab3}Ablation study of the effectiveness of our proposed components on THUMOS-14 \cite{jiang2015url}. CI and EE indicate additionally using the current information and early embedding input information, respectively.}
\end{table}

We conduct the same experiment on the THUMOS-14 dataset \cite{jiang2015url} to confirm the generality of the proposed components.
We obtain performance gains as individually incorporating the proposed components into GRU (see Table \ref{tab3}), where our IDN achieves an improvement of 3.3\% mAP compared to Baseline.
These results successfully demonstrate the effectiveness and generality of our components.

\begin{figure}[t!]
\centering{\includegraphics[width=.97\linewidth]{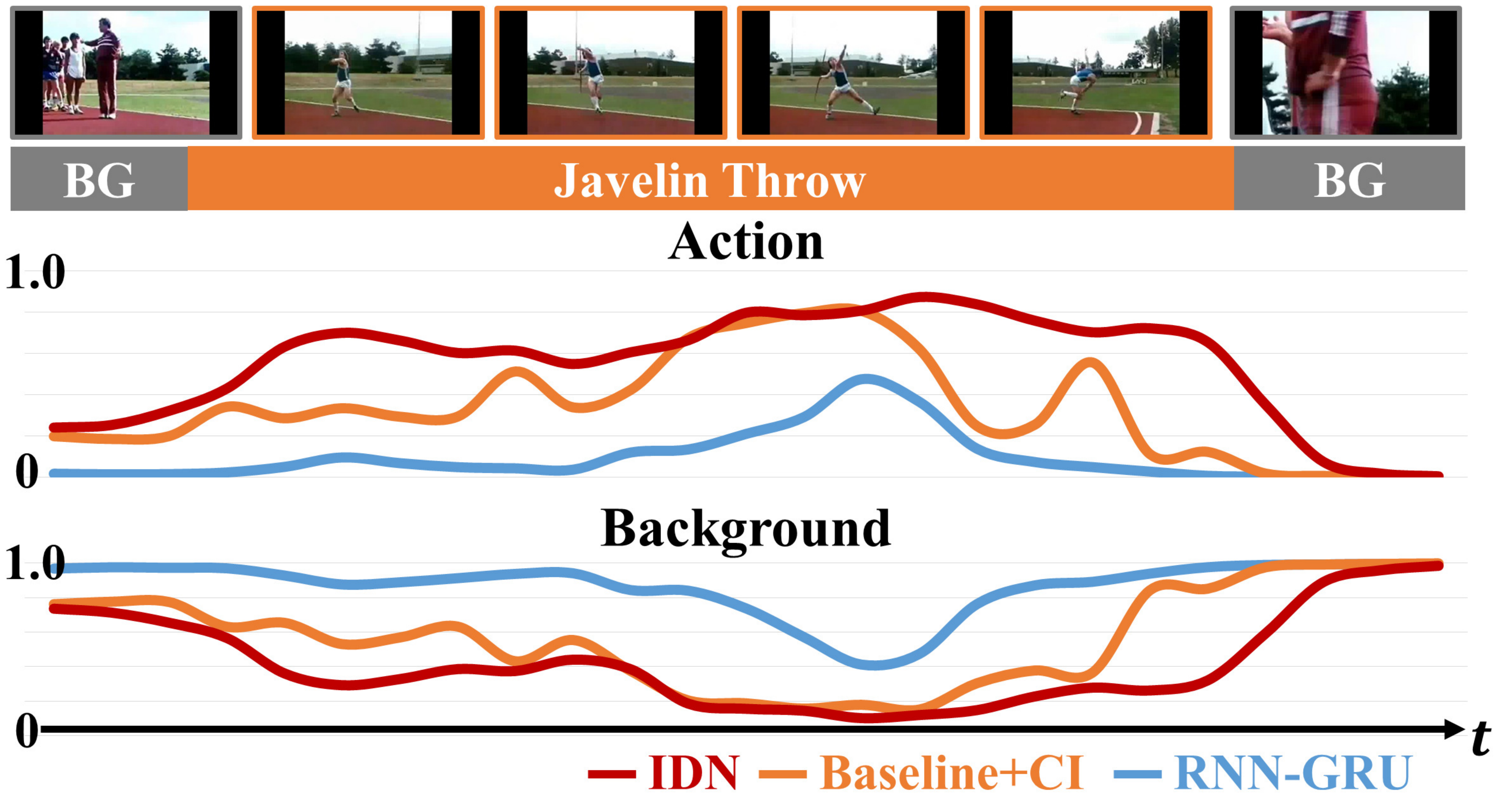}}
\centering{\caption{\label{fig6}Qualitative comparisons on predicted and GT probabilities for action (top) and background (bottom).}
}
\end{figure}

Figure \ref{fig6} shows qualitative comparisons on predicted and GT probabilities, where our IDN achieves the best results on both action and background frames.

\begin{figure}[t!]
\centering{\includegraphics[width=.97\linewidth]{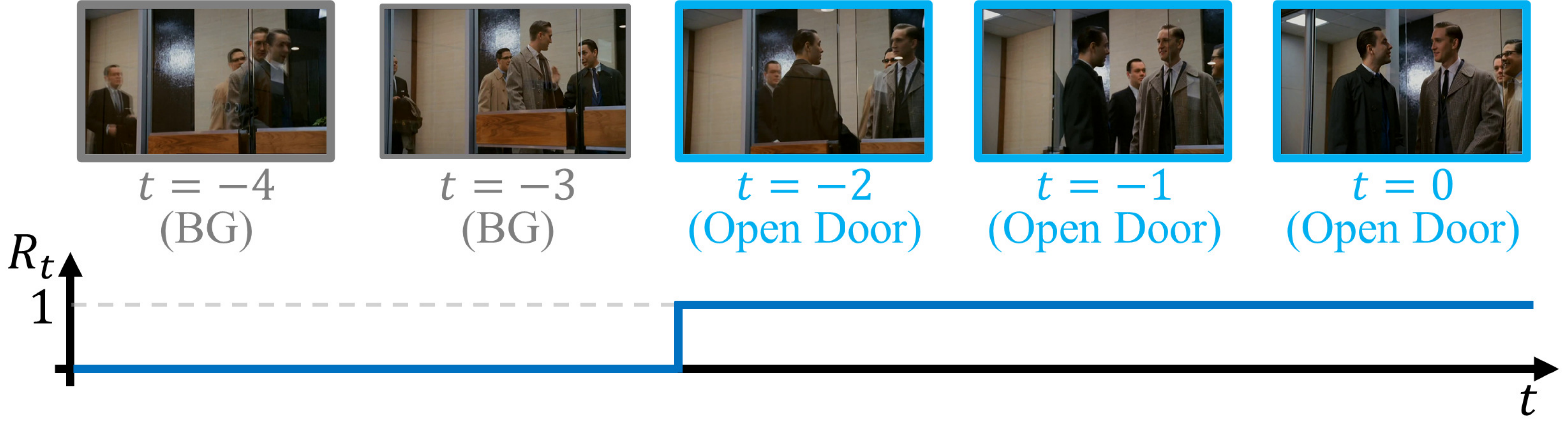}}
\centering{\caption{\label{fig3}Example of relevance scores $R_t$ of input chunks.}}
\end{figure}

\begin{figure}[t!]
\begin{minipage}[t]{0.99\linewidth}
\begin{center}
\includegraphics[width=.98\linewidth]{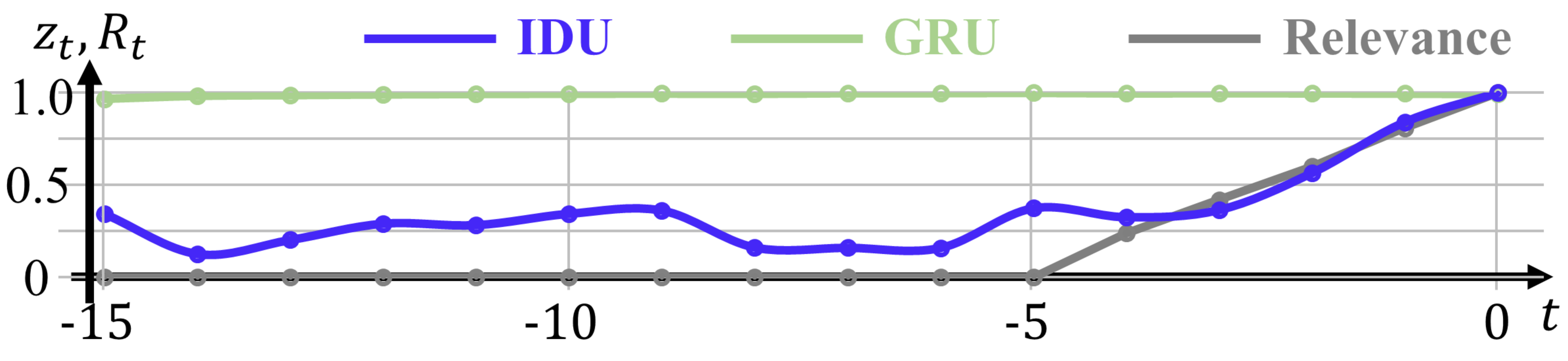}
\end{center}
\vspace{-0.4cm}
\small{(a) On the input sequences containing from one to five relevant chunks (i.e., from $t=0$ to $t=-4$). }
\end{minipage}
\hfill \vspace{0.1cm}
\begin{minipage}[t]{0.99\linewidth}
\begin{center}
\includegraphics[width=.98\linewidth]{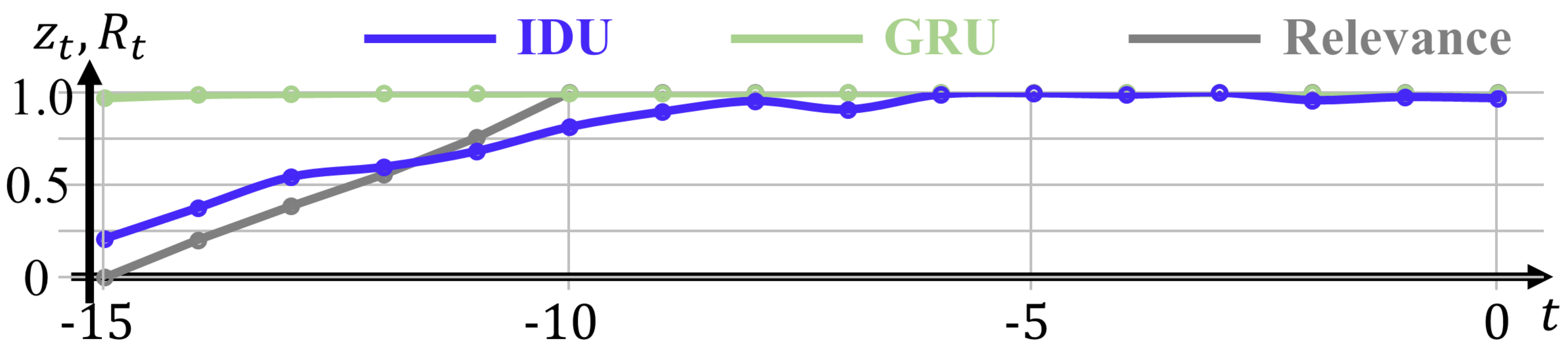}
\end{center}
\vspace{-0.4cm}
\small{(b) On the input sequences containing from 11 to 15 relevant chunks (i.e., from $t=-10$ to $t=-14$).}
\end{minipage}
\hfill \vspace{0.2cm}
\caption{\label{fig4}Comparison between the update gate $z_t$ values of our IDU and GRU \cite{cho2014emnlp}. Update gate values are measured on the input sequences containing (a) from one to five relevant chunks and (b) from 11 to 15 relevant chunks.}
\end{figure}

To confirm the effect of our components, we compare the values of the update gates $z_t$ between our IDU and GRU.
For a reference, we introduce the relevance score $R_t$ of each chunk regarding a current action.
Specifically, we set the scores of input chunks representing the current action as 1, otherwise 0 (see Fig. \ref{fig3}).
Note that the update gate controls how much information from the input will carry over to the hidden state.
Therefore, the update gate should drop the irrelevant information and pass over the relevant information related to the current action.
In Fig. \ref{fig4}, we plot the $z_t$ values of IDU and GRU and relevance scores against each time step.
On the input sequences containing from one to five relevant chunks, the $z_t$ values of GRU are very high at all time steps.
In contrast, our IDU successfully learns the $z_t$ values following the relevance scores (see Fig. \ref{fig4}(a)).
We also plot the average $z_t$ values on the input sequences including from 11 to 15 relevant chunks in Fig. \ref{fig4}(b), where our IDU yields the $z_t$ values similar to the relevance scores.
These results demonstrate that our IDU effectively models the relevance of input information to the ongoing action.

Compared to GRU, IDU has additional weights ${\bf{W}}_{xe}\in \mathbb{R}^{d_x \times 512}$ and ${\bf{W}}_{ep}\in \mathbb{R}^{512 \times (K+1)}$ in the early embedding module.
Our early embedding module reduces the dimensions of $x_t$, $x_0 \in \mathbb{R}^{d_x \times 512}$, which makes the parameters (i.e., ${\bf{W}}_{x_0r}$, ${\bf{W}}_{x_tz} \in \mathbb{R}^{512 \times 512}$) in IDU less than the parameters (i.e., ${\bf{W}}_{xr}$, ${\bf{W}}_{xz}\in \mathbb{R}^{d_x \times 512}$) in GRU.
The other weights have the same number of parameters in IDU and GRU.
As a result, the number of parameters for IDU is 75.3\% of that for GRU with $d_x=3072$ and $K=20$.
%We also confirmed that RNN-GRU and IDN operate 531fps and 526fps with a single GTX 1080Ti, respectively.}

\subsection{Performance Comparison}
In this section, we compare our IDN with state-of-the-art methods on TVSeries \cite{geest2016eccv} and THUMOS-14 \cite{jiang2015url} datasets.
We use three types of input including RGB, Flow, and Two-Stream.
As the input of our IDU, we take only appearance features for the RGB input and motion features for the Flow input.
IDN, TRN \cite{xu2019iccv}, RED \cite{gao22017bmvc}, and ED \cite{gao22017bmvc} use same two-stream features for the Two-Stream input, which allows a fair comparison.
We also employ another feature extractor, the TSN model \cite{wang2016eccv} pretrained on the Kinetics dataset \cite{carreira2017cvpr}.
We name our IDN with this feature extractor IDN-Kinetics.

%\begin{table}[t]
%\centering
%\begin{tabular}{C{2.3cm}|C{2.8cm}|C{1.5cm}}\hline 
%Input & Method & mcAP (\%) \\ \hline\hline
%\multirow{8}{*}{RGB}&CNN \cite{donahue2015cvpr} & 60.8 \\
%&LRCN \cite{donahue2015cvpr} & 64.1 \\
%&ED \cite{gao22017bmvc}		 & 71.0 \\
%&RED \cite{gao22017bmvc}		 & 71.2 \\
%&LSTM \cite{geest2018wacv} & 71.4 \\
%&2S-FN \cite{geest2018wacv} & 72.4 \\
%&TRN \cite{xu2019iccv} 		 & 75.4 \\
%&IDN      		 & \bf{76.6} \\ \hline
%\multirow{2}{*}{Flow}&FV-SVM \cite{geest2016eccv} & 74.3 \\ 
%&IDN 		 & \bf{80.3} \\ \hline
%\multirow{5}{*}{\shortstack{Two-Stream}}&ED \cite{gao22017bmvc} 		 & 78.5 \\
%&RED \cite{gao22017bmvc}		 & 79.2 \\
%&TRN \cite{xu2019iccv} 		 & 83.7 \\
%&IDN      		 & \bf{84.7} \\ \cdashline{2-3}
%&IDN-Kinetics      		 & \bf{86.1} \\ \hline
%\end{tabular}
%\centering\caption{\label{tab4}Performance comparison on TVSeries \cite{geest2016eccv}. All methods except LRCN \cite{donahue2015cvpr} and CNN \cite{donahue2015cvpr} are proposed for online action detection. IDN, TRN \cite{xu2019iccv}, RED \cite{gao22017bmvc}, and ED \cite{gao22017bmvc} use same two-stream features for the Two-Stream input.}
%\end{table}

\begin{table}[t]
\centering
\begin{tabular}{C{2.5cm}|C{3.cm}|C{1.5cm}}\hline 
Input & Method & mcAP (\%) \\ \hline\hline
\multirow{5}{*}{RGB}&LRCN \cite{donahue2015cvpr} & 64.1 \\
&RED \cite{gao22017bmvc}		 & 71.2 \\
&2S-FN \cite{geest2018wacv} & 72.4 \\
&TRN \cite{xu2019iccv} 		 & 75.4 \\
&IDN      		 & \bf{76.6} \\ \hline
\multirow{2}{*}{Flow}&FV-SVM \cite{geest2016eccv} & 74.3 \\ 
&IDN 		 & \bf{80.3} \\ \hline
\multirow{4}{*}{\shortstack{Two-Stream}}&RED \cite{gao22017bmvc}		 & 79.2 \\
&TRN \cite{xu2019iccv} 		 & 83.7 \\
&IDN      		 & \bf{84.7} \\ \cdashline{2-3}
&IDN-Kinetics      		 & \bf{86.1} \\ \hline
\end{tabular}
\centering\caption{\label{tab4}Performance comparison on TVSeries \cite{geest2016eccv}. IDN, TRN \cite{xu2019iccv}, RED \cite{gao22017bmvc}, and ED \cite{gao22017bmvc} use same two-stream features for the Two-Stream input.}
\end{table}

\begin{table}[t]
\centering
\begin{tabular}{C{2.5cm}|C{3.cm}|C{1.5cm}}\hline 
Setting & Method & mAP (\%) \\ \hline\hline
\multirow{5}{*}{Offline} & CNN \cite{simonyan2015iclr} & 34.7 \\
 & CNN \cite{simonyan2014nips} & 36.2 \\
 & LRCN \cite{donahue2015cvpr}  & 39.3 \\
 & MultiLSTM \cite{yeung2018ijcv} & 41.3 \\
 & CDC \cite{shou2017cvpr} & 44.4 \\ \hline
\multirow{4}{*}{Online}  & RED \cite{gao22017bmvc} & 45.3 \\
%&ED \cite{gao22017bmvc} 		 & 43.7 \\
 & TRN \cite{xu2019iccv} 		 & 47.2 \\
 & IDN		      		 & \bf{50.0} \\ \cdashline{2-3}
 & IDN-Kinetics      		 & \bf{60.3} \\ \hline
\end{tabular}
\centering\caption{\label{tab5}Performance comparison on THUMOS-14 \cite{jiang2015url}. IDN, TRN \cite{xu2019iccv}, RED \cite{gao22017bmvc}, and ED \cite{gao22017bmvc} use same two-stream features.}
\end{table}

We report the results on TVSeries in Table \ref{tab4}.
Our IDN significantly outperforms state-of-the-art methods on all types of input, where IDN achieves 76.6\% mcAP on the RGB input, 80.3\% mcAP on the Flow input, and 84.1\% mcAP on the Two-Stream input.
Furthermore, IDN-Kinetics achieves the best performance of 86.1\% mcAP.
Note that IDN effectively reduces wrong detection results occurred from the irrelevant information by discriminating the relevant information.
However, 2S-FN, RED, and TRN accumulates the input information without considering its relevance to an ongoing action.
In addition, our IDN yields better performance than TRN \cite{xu2019iccv} although IDN takes shorter temporal information than IDN (i.e., 16 chunks vs. 64 chunks).

\begin{table*}[t]
\centering
\begin{tabular}{C{4.5cm} C{.8cm} C{.8cm} C{.8cm} C{.8cm} C{.8cm} C{.8cm} C{.8cm} C{.8cm} C{.8cm} C{.8cm}}\hline 
& \multicolumn{10}{c}{Portion of action} \\ \cline{2-11}
Method & 0\%-10\% & 10\%-20\% & 20\%-30\% & 30\%-40\% & 40\%-50\% & 50\%-60\% & 60\%-70\% & 70\%-80\% & 80\%-90\% & 90\%-100\% \\ \hline\hline
\multicolumn{1}{c|}{CNN \cite{geest2016eccv}} & 61.0 & 61.0 & 61.2 & 61.1 & 61.2 & 61.2 & 61.3 & 61.5 & 61.4 & 61.5 \\
\multicolumn{1}{c|}{LSTM \cite{geest2016eccv}} & 63.3 & 64.5 & 64.5 & 64.3 & 65.0 & 64.7 & 64.4 & 64.4 & 64.4 & 64.3 \\
\multicolumn{1}{c|}{FV-SVM \cite{geest2016eccv}} & 67.0 & 68.4 & 69.9 & 71.3 & 73.0 & 74.0 & 75.0 & 75.4 & 76.5 & 76.8 \\
\multicolumn{1}{c|}{TRN \cite{xu2019iccv}} & 78.8 & 79.6 & 80.4 & 81.0 & 81.6 & 81.9 & 82.3 & 82.7 & 82.9 & 83.3 \\
\multicolumn{1}{c|}{IDN}  & 80.6 & 81.1 & 81.9 & 82.3 & 82.6 & 82.8 & 82.6 & 82.9 & 83.0 & 83.9 \\ 
\multicolumn{1}{c|}{IDN-Kinetics}  & \bf{81.7} & \bf{81.9} & \bf{83.1} & \bf{82.9} & \bf{83.2} & \bf{83.2} & \bf{83.2} & \bf{83.0} & \bf{83.3} & \bf{86.6} \\ \hline
\end{tabular}
\caption{\label{tab6}Performance comparison for different portions of actions on TVSeries \cite{geest2016eccv} in terms of mcAP (\%). The corresponding portions of actions are only used to compute mcAP after detecting current actions on all frames in an online manner.}
\end{table*}

\begin{figure*}[t!]
\begin{center}
{\centering{\includegraphics[width=.99\linewidth]{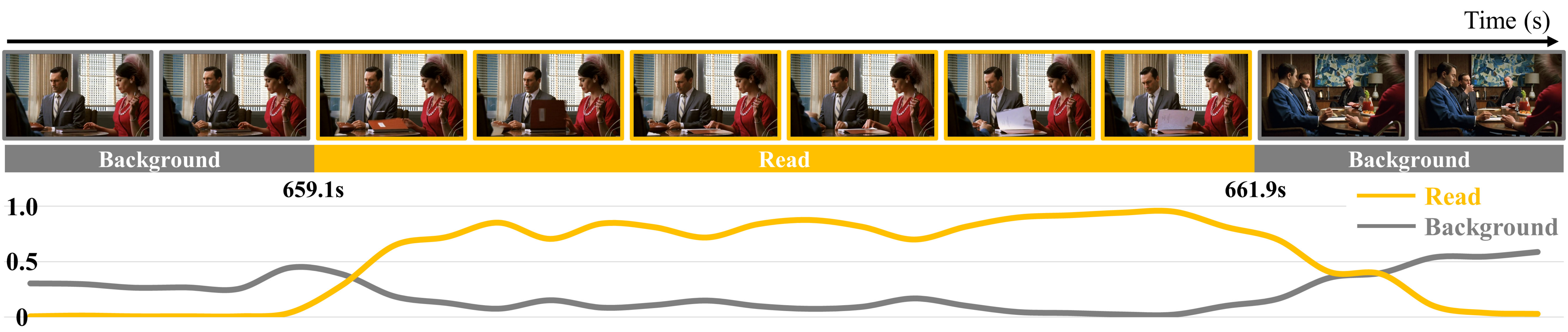}}\vspace{0.1cm}}
{\centering{\includegraphics[width=.99\linewidth]{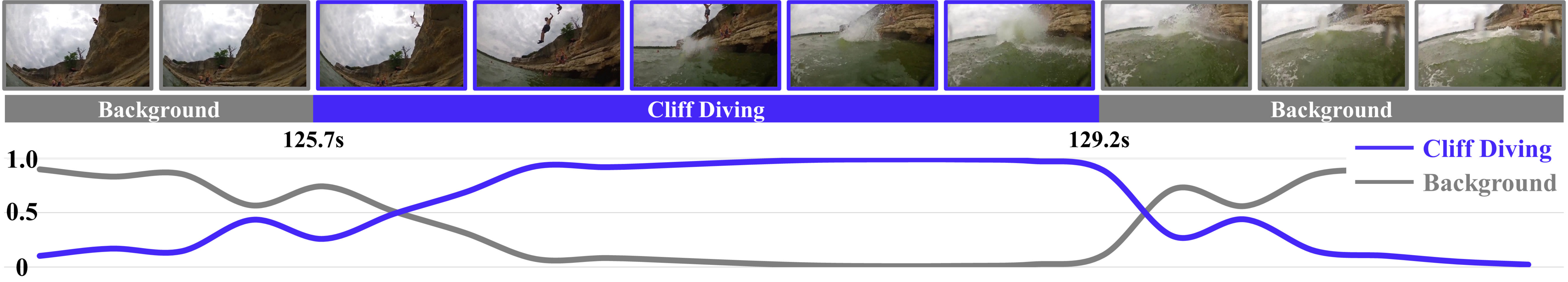}}}
\vfill \vspace{-0.3cm}
\end{center}
\caption{\label{fig5}Qualitative evaluation of IDN on TVSeries \cite{geest2016eccv} (top) and THUMOS-14 \cite{jiang2015url} (bottom). Each result shows frames, ground truth, and estimated probabilities.}
\end{figure*}

In Table \ref{tab5}, we compare performances between our IDN and state-of-the-art approaches for online and offline action detection.
The compared offline action detection methods perform frame-level prediction.
As a result, both IDN and IDN-Kinetics outperforms all methods by a large margin.

In online action detection, it is important to identify actions as early as possible.
To compare this ability, we measure the mcAP values for every 10\% portion of actions on TVSeries.
Table \ref{tab6} shows the comparison results among IDN, IDN-Kinetics, and previous methods, where our methods achieve state-of-the-art performance at every time interval.
This demonstrates the superiority of our IDU in identifying actions at early stages as well as all stages.

\subsection{Qualitative Evaluation}
For qualitative evaluation, we visualize our results on TVSeries \cite{geest2016eccv} and THUMOS-14 \cite{jiang2015url} in Fig. \ref{fig5}.
The results on the TVSeries dataset show high probabilities on the true action label and reliable start and end time points.
Note that identifying actions at the early stage is very challenging in this scene because the only subtle change (i.e., opening a book) happens.
On THUMOS-14, our IDN successfully identifies ongoing actions by yielding the contrasting probabilities between true action and background labels.

\section{Conclusion}
In this paper, we proposed IDU that extends GRU \cite{cho2014emnlp} with two novel components: 1) a mechanism using current information and 2) an early embedding module.
These components enable IDU to effectively decide whether input information is relevant to a current action at every time step.
Based on IDU, our IDN effectively learns to discriminate relevant information from irrelevant information for identifying ongoing actions.
In comprehensive ablation studies, we demonstrated the generality and effectiveness of our proposed components.
Moreover, we confirmed that our IDN significantly outperforms state-of-the-art methods on TVSeries \cite{geest2016eccv} and THUMOS-14 \cite{jiang2015url} datasets for online action detection.

\section*{Acknowledgments}
This work was supported by Institute of Information \& Communications Technology Planning \& Evaluation (IITP) grant funded by the Korea government (MSIT) (No.B0101-15-0266, Development of High Performance Visual BigData Discovery Platform for Large-Scale Realtime Data Analysis).

\newpage

{\small
\bibliographystyle{ieee_fullname}
\bibliography{mybibfile}
}

\end{document}